\documentclass[letterpaper, 10 pt, conference]{ieeeconf}  

\IEEEoverridecommandlockouts                              

\usepackage{graphics} 
\usepackage{graphicx}

\usepackage{epsfig} 
\usepackage{amsmath} 
\usepackage{amssymb}  

\usepackage{algorithm}
\usepackage[noend]{algpseudocode} 
\algnewcommand\AAND{\textbf{ and }}
\algnewcommand\Or{\textbf{ or }}

\usepackage{color}
\usepackage{citesort}
\usepackage{url}
\usepackage{breakurl}
\usepackage[breaklinks]{hyperref}

\DeclareMathAlphabet{\pazocal}{OMS}{zplm}{m}{n}

\DeclareMathAlphabet{\mathpzc}{OT1}{pzc}{m}{it}

\usepackage{array}
\newcolumntype{C}[1]{>{\centering\arraybackslash}p{#1}}
\newcolumntype{M}[1]{>{\raggedright\arraybackslash}p{#1}}

\usepackage{array} 
\newcolumntype{L}[1]{>{\raggedright\let\newline\\\arraybackslash\hspace{0pt}}m{#1}}	
\newcolumntype{S}[1]{>{\centering\let\newline\\\arraybackslash\hspace{0pt}}m{#1}}
\newcolumntype{R}[1]{>{\raggedleft\let\newline\\\arraybackslash\hspace{0pt}}m{#1}}

\def\*#1{\mathbf{#1}}
\def\@#1{\boldsymbol{#1}}

\makeatletter
\renewcommand*{\@opargbegintheorem}[3]{\trivlist
  \item[\hskip \labelsep{\itshape #1\ #2}] \textit{(#3)}\ }
\makeatother

\title{\LARGE \bf
Collision-tolerant Aerial Robots: A Survey
}

\author{Paolo De Petris$^1$, Stephen J. Carlson$^2$, Christos Papachristos$^2$, and Kostas Alexis$^1$
\thanks{This material was supported by the Research Council of Norway under project SENTIENT (grant number 321435).}
\thanks{$^1$ The authors are with the Autonomous Robots Lab, Norwegian University of Science and Technology (NTNU), O. S. Bragstads Plass 2D, 7034, Trondheim, Norway {\tt\small paolo.de.petris@ntnu.no}}%
\thanks{$^2$ The authors are with the Robotic Workers Lab, University of Nevada, Reno (UNR), 1664 N. Virginia St., 89557, Reno, Nevada, USA {\tt\small stephen\_carlson@nevada.unr.edu}}%
}

\begin{document}

\maketitle
\thispagestyle{empty}
\pagestyle{empty}

\begin{abstract}
As aerial robots are tasked to navigate environments of increased complexity, embedding collision tolerance in their design becomes important. In this survey we review the current state-of-the-art within the niche field of collision-tolerant micro aerial vehicles and present different design approaches identified in the literature, as well as methods that have focused on autonomy functionalities that exploit collision resilience. Subsequently, we discuss the relevance to biological systems and provide our view on key directions of future fruitful research. 
\end{abstract}

\section{INTRODUCTION}

Aerial robots are being utilized in an ever increasing set of application domains. Among others, those that involve challenging and cluttered environments or the interaction with people are the most demanding regarding the safety of the system and of its surroundings. Accordingly, vast resources are put into the effort of developing robust and high-performance collision avoidance systems. However, a collision may not always be definitely avoidable and this can be demonstrated both theoretically~\cite{karaman2012high} and by observing how insects and birds fly as they often experience impact with their environment or among themselves~\cite{ma2015function,mountcastle2014biomechanical,henningsson2021flying,hudson1936studies,rubenson2022running}. Responding to this fact and focusing specifically on small-scale systems, the field of Collision-Tolerant Micro Aerial Vehicles (CT-MAVs) has been developed. In this survey we review the different designs for collision-tolerant aerial robots and present notable features of several solutions. Among others we discuss rigid, elastic, origami-like, tensegrity-based, gimbal-based, bioinspired, expandable, morphing or foldable, bimodal aerial/ground, and reconfigurable collision resilient flying robot designs. As the majority of the systems are at their core a quadrotor or a multirotor design, we explicitly mention it when this is not the case. 

A generalized definition of Collision-Tolerance in small aerial robots is drawn given the behaviors of a) ``Crash-resistance'', b) ``Contact-resistance'', and c) ``Upset-tolerance'': 

\begin{itemize}
  \item Crash-resistance is the ability of a flying robot to survive forcible impacts with environment elements and the associated release of kinetic energy. If a design can consistently survive blunt collisions near-to or exceeding the typical cruise velocity of the vehicle, the system can be defined as being crash-resistant, and the kinetic energy at which this occurs can be used as a metric for comparison between designs (at least between systems of similar scale). 
  \item Contact-resistance is the tendency for the design to reject or cope with passive or incidental contact with the environment, with collision energies well below the Crash-Resistant metric defined above. Any design that protects the robot's propulsion system or that can operate while making continuous smooth contact with environmental hazards can be considered Contact-Resistant.
  \item Upset-tolerance is the ability for the design to self-right in the event that a collision event causes the vehicle to depart from flight and come to rest in a pose or position in the environment from which it would not typically enter or resume flight. This is the ability to recover from being flipped upside-down, or to relocate from an obstructed position using alternative locomotion methods.
\end{itemize}
Beyond the above three properties that directly refer to the collision-tolerance capabilities of a flying robot, two further properties are of significant value for this study as they influence how a system may experience interaction, namely ``softness'', ``fold-ability'', and ``squeeze-ability'' as discussed below:
\begin{itemize}
  \item Softness refers to the ability of the robotic embodiment to present elasticity and dampen the effects of the kinetic energy released during impact. Most designs are typically very stiff, but a set of elastic designs have recently emerged.
  \item Fold-ability refers to the ability of the embodiment to passively or actively fold and reconfigure its shape thus modifying its cross-section, which in turn may allow to either avoid the collision or change the geometry of interaction.
  \item Squeeze-ability refers to the ability of the flying robot to squeeze its body when encountering an impact and especially as this relates to traversing narrow passages.
\end{itemize}


The remainder of this paper is organized as follows. Section~\ref{sec:review} reviews the current state of the art regarding the design of Collision-Tolerant MAVs, followed by Section~\ref{sec:autonomy} that identifies works that have focused on the interplay between collision resilience and autonomy. We discuss certain observations from studies in nature and their relevance to aerial robotic collision-tolerant designs in Section~\ref{sec:discussion}, before we outline directions for future research in Section~\ref{sec:future}. Finally, conclusions are drawn in Section~\ref{sec:concl}.

\section{A TAXONOMY OF CT-MAVs}\label{sec:review}

\begin{figure*}[ht!]
\includegraphics[width=\textwidth]{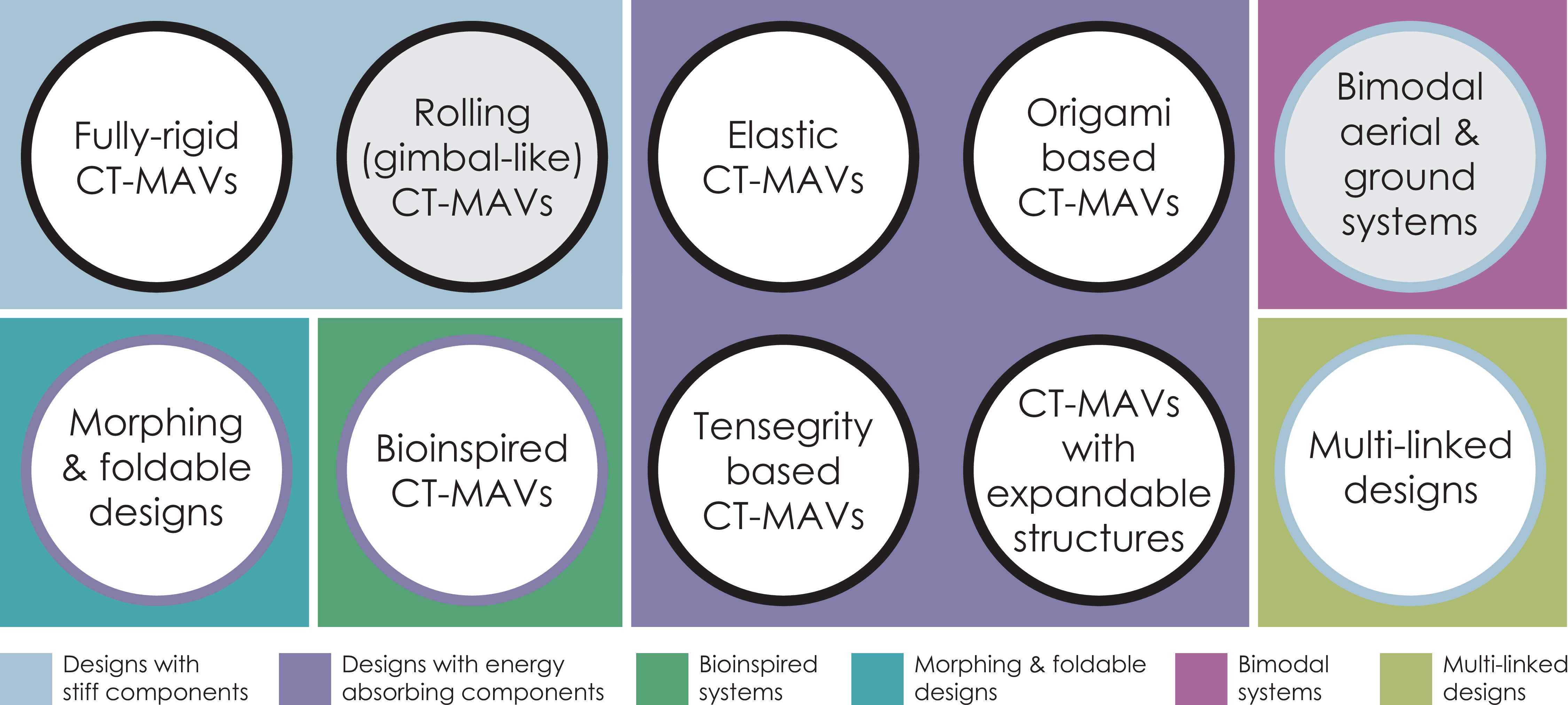}
\caption{A loose taxonomy of collision-tolerant micro aerial vehicles. Different background colors are used for different core categories of design such as those with only stiff components and those specifically involving energy-absorbing elements. Bioinspired systems and morphing/foldable design goals exceed collision tolerance but importantly also often involve compliant components thus their outer circle is depicted with the background color of those designs involving energy-absorbing components and being explicitly focused to collision-tolerance. Bimodal and multi-linked designs also have their own design goals exceeding collision-tolerance but at least in the literature presented in this review tend to employ all-stiff components (outer circle color) although this is not always the case. Bimodal aerial and ground robots often present rolling cages which often makes them particularly relevant to the rolling (gimbal-like) CT-MAVs (thus sharing a common circle color).  }
\vspace{-1ex}
\label{fig:ctmavstaxonomy}
\end{figure*}

Depending on the design approach and the core research focus relevant to collision tolerance, the following CT-MAVs sub-categories are presented in the literature: a) fully-rigid, stiff, designs without additional moving parts, b) general elastic or compliant designs that present softness and possibly the ability to squeeze, c) origami-based designs that accordingly may also present softness, d) tensegrity-based vehicles also presenting softness, e) gimbal-based systems, f) bioinspired approaches, g) CT-MAVs with expandable structures, h) morphing or foldable robots, i) bimodal aerial/ground designs, as well as j) multi-linked multi-system designs. Within those sub-categories certain taxonomies can be identified. The elastic, origami-based and tensegrity-based designs can be thought as belonging to a broader category of systems that present softness and involve energy-absorbing mechanisms to reduce the effects of impact towards certain parts of the robot. Bioinspired systems also tend to involve compliant components. Partially, this also holds for morphing or foldable robots. Bimodal and multi-linked systems represent their own individual subcategories, although certain bimodal designs have commonalities with vehicles in the gimbal-based category and others. Subsequently, we outline key findings within these categories. It is noted that in certain cases, a design might belong to more than one category. In this case we mostly include it in the category of highest relevance but the description should allow to understand its fit to other design approaches too. Figure~\ref{fig:ctmavstaxonomy} outlines a loose taxonomy of CT-MAVs and relevant designs. 

\subsection{Rigid/Stiff CT-MAVs without further moving parts}

The most straightforward design approach to realize collision-tolerance on Micro Aerial Vehicles is through the utilization of rigid protective shrouds. Equipping multirotor systems with protective shrouds is nothing new~\cite{hoffmann2007quadrotor,hoffmann2011precision}. Simultaneously, several designs implement implicitly protection around the most significant moving parts, the propellers, including that of ducted fan systems~\cite{marconi2012control}. In this section, we focus on systems and contributions that explicitly address their collision-tolerance properties or demonstrate how they can benefit from those. The authors in~\cite{wang2020fly} presented a framework for sensor-based reactive collision detection, estimation, and recovery, and simultaneously proposed a potential fields-based method to map the collision in order to derive an appropriate maneuver to avoid the collided object and ensure the safety of the system. The work in~\cite{mulgaonkar2015design} presented a miniaturize-sized collision-tolerant quadrotor design that is capable of sustaining significant collisions and recover from those, while in~\cite{mulgaonkar2017robust} the design is exploited jointly with a simple motion planner able to swarm through complex environments. The system is extremely lightweight (25g system weight with a surrounding frame of 3g) and the results included collisions with speeds up to $4\textrm{m/s}$. Extending upon these exciting results, the authors in~\cite{mulgaonkar2020tiercel} presented ``TIERCEL'', a novel CT-MAV that implements a rigid collision-tolerant frame, integrates vision-based localization and mapping capabilities while being particularly lightweight, exploits contact to deal with transparent or reflective surfaces, and demonstrates an impressive thrust-to-weight ratio of $6$:$1$. 

Focusing on advanced autonomy and integrating all the essential components of the software stack of CERBERUS~\cite{tranzatto2022cerberus,tranzatto2022team,tranzatto2022cerberus,cerberus_phaseI_phaseII}, the winning team of the DARPA Subterranean Challenge, ``RMF-Owl'' in~\cite{rmfowl_icuas,kulkarni2022autonomous} presents a collision-tolerant frame on a robot with a total weight just below $1.5\textrm{kg}$, a total airframe weight of $145\textrm{g}$, and a thrust-to-weight ratio of $2$:$1$ despite the integration of an OUSTER OS0/OS1 3D LiDAR, cameras, advanced compute, onboard lights and more. A modified version of the system~\cite{nguyen2022motion} integrates a Realsense D455, a Realsense T265, and an NVIDIA Xavier NX CPU/GPU board and relies on a deep learned collision predictor for safe environment navigation, while collision-tolerance is considered as a mean to mitigate the effects of an inadvertently occurring collision. An earlier and smaller version of the system, only relying on visual-inertial localization, was utilized for the purposes of radiation monitoring and spectroscopic characterization~\cite{mascarich2021autonomous}. The particular airframe was originally designed having also compliant components and is discussed in the next subsection. Finally, it is noted that a set of other designs have been proposed including vehicles with active control surfaces~\cite{moon2018uni}, as well as cubical frames for omnidirectional collision-tolerant MAVs~\cite{brescianini2018omni}. 


\subsection{Elastic CT-MAVs}

One of the most significant sub-categories of CT-MAVs corresponds to elastic/compliant designs that allow for embodiment softness. As opposed to fully rigid/stiff systems, elastic designs can provide enhanced performance in terms of damping collision forces and mitigating the effects of impact. The authors in~\cite{de2021flexible,de2021being} present a soft aerial robot inspired by the exoskeletons of arthropods. It fuses the protective cage and the main frame in one semi-rigid design with soft joints, reports a weight of less than $250\textrm{g}$, and demonstrated safe collisions with speeds up to $7\textrm{m/s}$. The work in~\cite{de2021resilient} presented the first version of RMF, a quadrotor with a rigid protective frame that further integrated compliant flaps and focused on an autonomy stack that allowed it to navigate through extremely tight, manhole-sized, environments. 

The set of authors in~\cite{klaptocz2012design,klaptocz2013euler,briod2012airburr} presented multiple CT-MAV systems with their most notable feature being the utilization of Euler spring-based compliant protective designs. One of the presented systems realizes a protective structure in the form of Euler springs in a tetrahedral configuration such that the embedded elasticity best absorbs the energy of an impact, while simultaneously reducing the effect of forces acting on the robot's stiff inner frame. Extending upon the principle of elasticity-based mitigation of collisions, the work in~\cite{mintchev2017insect} presented a quadrotor with high survivability against collisions. In particular, it implements a dual-stiffness mechanism as a frame that is connected to the inner core (which houses the main navigation electronics) with magnets. The elements contributing to collision-tolerance are magnets, foam inserts, and elastic bands, and have a sum weight of $4.66\textrm{g}$ which represents $9\%$ of the total weight of the robot. The system demonstrated safe drop from significant heights.

Employing a relatively more traditional quadrotor design, the authors in~\cite{liu2021toward} introduce an actively-resilient design which features a compliant arm to absorb impact shocks, alongside a collision detection and characterization process and an associated recovery control algorithm generating smooth post-collision trajectories. The system relies on motion capture for the presented experiments --and does not incorporate sensing that would allow for Simultaneous Localization And Mapping (SLAM) or other such functionality-- but presents a high thrust-to-weight ratio ($4.58:1$) and reports safe collisions with speeds up to $5.9\textrm{m/s}$. 

Having a focus that goes beyond collision tolerance, the work in~\cite{patnaik2020design} presented ``SQUEEZE'', a novel quadrotor that integrates a passive folding mechanism for its arms and protective shrouds. Its design not only allows it to mitigate the risks of collisions but also enables it to pass through very narrow passages by passively varying its geometry. Accordingly, the system presents the key feature of squeeze-ability which turns out to be essential when flying through highly cluttered or narrow settings. Albeit implemented in a different manner, squeeze-ability is also presented by SquAshy~\cite{fabris2021soft,fabris2022crash}, a multi-modal system tailored to the exploration of confined environments. 

Whereas the above designs implement compliance on the core airframe, the contribution in~\cite{bui2022tombo} investigated the potential of soft, self-recovering, propellers. The authors present ``Tombo'' which is a propeller that can deform upon collision and recover to then allow the robot to continue its normal operation. The work further presents its aerodynamic model and other key design properties. Considering larger fixed-wing systems, the contributions in~\cite{cadogan2004morphing,cadogan2003inflatable} have considered the potential of inflatable wings both as a means for low-volume packaging and as a means of compliance for enhanced survivability (e.g., during possible impact at landing).

Employing a fundamentally different approach, the work in~\cite{chen2021collision} employs a micro-sized insect-like design and showed that soft-actuated subgram MAVs can demonstrate particularly robust and agile flight performance. It has a lift-to-weight ratio of $2.2$:$1$ and a total weight of $665\textrm{mg}$. Overall, we comment on the difficulty of designing efficient soft MAVs - a process in which designs with anisotropic compliance is important.


\subsection{Origami-based CT-MAVs}

A new approach to design CT-MAVs has recently emerged and involves the utilization of origami-based structures. Among the most notable designs, the ``Miura-oRing'' spinning cycle system and the Rotorigami, as presented in~\cite{sareh2018rotorigami,sareh2018spinning}, offer a lightweight and low-cost mechanical solution for impact protection around miniature-sized multirotor vehicles. The authors present the possible benefits of such systems with the Rotorigami demonstrating safe interaction with relatively significant speeds (e.g., $2\textrm{m/s}$). Importantly, origami allows to embed softness into the robot and at the same enables the utilization of a wide palette of materials, including high-performance engineering options such as carbon fiber. 

Earlier work in the domain has presented an origami-inspired cargo aerial robot~\cite{kornatowski2017origami}. The system implemented a safe and foldable design which in turn allows to shield the propellers when flying but simultaneously reduce its storage volume by $92\%$ when folded. 

Focusing on particularly miniaturized designs, the work in~\cite{shu2019quadrotor} proposed a passive foldable airframe that relies on origami principles for its design. The robot has a total mass of $51.2\textrm{g}$ with the weight of the foldable airframe being only $14.9\textrm{g}$. Furthermore, the work in~\cite{dilaverouglu2020minicore} presented an origami-based foldable quadrotor with its design allowing to safely mitigate the effects of collisions at low speeds. 

Of particular relevance to this domain of work is the study in~\cite{phan2020mechanisms}, where the mechanisms of collision recovery in flying beetles are studied. Looking specifically at the behavior of the larger hindwings of beetles during collision they found that origami-like folds within them could rapidly collapse on impact and bounce back afterwards. This allows them to act both as shock absorbers and stabilizers. This is important as this particular type of wing of beetles does not really flex, as opposed to other types of insect wings. 

\subsection{Tensegrity-based CT-MAVs}

An alternative approach to implementing an elastic collision-tolerant MAV design has emerged through the use of tensegrity structures. The authors in~\cite{zha2020collision} present one such system which specifically adopts the advantages of icosahedron tensegrity designs. The work presents a stress analysis for this tensegrity frame under collision forces and thus enables its optimization, alongside a specialized controller that allows to re-orient the vehicle while on the ground at any orientation. The system - as presented in this work - navigates by relying on motion capture, but presents a thrust-to-weight ratio as high as $3.4$:$1$ and demonstrated safe collisions with speeds up to $6.5\textrm{m/s}$. Following up, the work in~\cite{zha2022design} employs a model-based methodology to design a tensegrity collision-tolerant aerial robot. The methodology predicts the stresses in the structure so as to select components appropriately, while a re-orientation controller is also implemented to enable mission continuation post-collision. Additional designs involving tensegrity structures have been proposed~\cite{johnson2021feasibility}, while at the same time methods that exploit the interaction potential of tensegrity-based CT-MAVs for motion planning are also being investigated~\cite{savin2022mixed}. The work in~\cite{zappetti2022dual} presents broader results on dual stiffness tensegrity platforms for robotic systems. 

\subsection{Gimbal-based CT-MAVs}

A class of collision-tolerant MAVs with significant commercial success is that of vehicles implementing a gimbal-like inner stabilizing frame. The contribution in~\cite{briod2014collision} implements a spherical protective outer frame and an inner stabilizing gimbal system, and represents one of the most successful designs with commercial products associated with it having significant industry adoption. The system not only mitigates the structural integrity effects of impact but also allows for enhanced attitude stability during and after the collision. The particular design has a total weight of $385\textrm{g}$ with the protective frame weighing overall $84\textrm{g}$ and the gimbal mechanism $56\textrm{g}$. Other works have followed upon the design success of such a gimbal-based system~\cite{mizutani2015proposal,salaan2018close}. The commercial success of such a design is also reflected on the fact that it is found in specialized application-driven contributions on topics such as meshing and insulation strategies when such caged MAVs are to be used in proximal inspection of distribution utility lines~\cite{librado2022meshing}. 



Following upon the success of such systems, further designs have been proposed. Among others, another spherical design with a gimbal-like inner frame has been presented in~\cite{ramos2019spherical} and focused on implementing a complete navigation suite. From a different perspective, the work in~\cite{salaan2019development} implemented protective rotating spherical shells around each individual propeller of a quadrotor system, thus allowing to reduce the overall volume of the design. A related design is presented in~\cite{kalantari2020drivocopter} but utilizes actuated spherical wheels and is made for hybrid aerial/ground navigation capabilities.

\subsection{Bioinspired CT-MAVs}

Bioinspired designs are an upcoming domain in unmanned aerial vehicles research. Interestingly, a set of contributions specifically addresses the potential of bioinspired MAV systems to also offer collision-tolerance. Referenced earlier in this manuscript for its elastic collision-tolerance properties, the work in~\cite{chen2021collision} presents a sub-gram insect-like MAV that utilizes novel Dielectric Elastomer Actuators (DEA) which achieves high power density ($1.2\textrm{kW/kg}$) and significant transportation efficiency ($37\%$), thus allowing to deliver a high lift-to-weight ratio ($2.2$:$1$) and agile maneuverability. The robot demonstrates in-flight collision recovery and performs a somersault within just $0.16\textrm{s}$. 

The above insect-like system employs elastic components and is linked to a widespread trend in biological systems to utilize elastic components in their bodies~\cite{xing2020stiffness,ma2015function}. Yet other approaches also exist. Observing the hindwings in free flying rhinoceros beetles, the researchers in~\cite{phan2020mechanisms}, identified that although they cannot readily flex, they do implement folding-and-unfolding mechanisms based on origami-like structures thus acting as absorbers and stabilizers. This behavior was then reflected on a newly designed flapping-wing MAV able to afford collisions at low speeds. 

Considering flapping wings, the work in~\cite{tu2021flying} evaluates the disturbing effects of wing damage --often due to collisions-- on flight, as lift gets significantly reduced and aerodynamic damping varies as wing areas change. It also proposes a controller to cope with such detrimental effects thus maximizing the system's survivability. Furthermore, the authors in~\cite{sihite2020computational} propose an armwing mechanism motivated by biological studies and present results that can contribute, among others, to the design of winged MAVs with enhanced collision-tolerance.

Focusing on MAV systems that can present efficient fixed-wing forward flight but also the multi-modality of terrestrial locomotion, the work in~\cite{vourtsis2021robotic} presents the ``Robotic Elytra''. The design of the system is motivated by the need to enable collision-tolerance for forward-flying robots without major negative impact on their aerodynamic properties due to the drag that a cage would induce. Motivated by the wing system found in beetles, it adds a set of retractable wings that can rapidly encapsulate the main folding wings when protection is necessary. The robot presents dual-navigation capabilities, namely both flying and terrestrial locomotion.

The above are examples of designs that are particularly bioinspired. Simultaneously, a set of designs exist where they get biological motivation and then design around a rather more traditional human engineering-driven solution~\cite{mulgaonkar2017robust,mintchev2017insect}.

\subsection{CT-MAVs with Expandable Structures}

The majority of CT-MAVs present a protecting structure around their core airframe and/or elasticity within it. A different approach is possible and involves the design of expandable protective structures. The authors in~\cite{hedayati2020pufferbot} presented ``PufferBot'' which is based around a quadrotor integrating an expandable structure that may selectively expand in order to protect the vehicle's propellers when in close proximity with obstacles in the environment. The expanding element is located on top of the vehicle and is based on a 3D-printed expandable scissor structure combined with a one degree-of-freedom actuator with a rack and pinion mechanism. Four different designs of this expanding structure are presented and address different interaction conditions. The robot can handle safely collisions with forces in the range of $6$-$9\textrm{N}$. 

Relevant to the case of expandable structures, a morphing cargo drone presented in~\cite{kornatowski2020morphing} has two modes of a protective cage, namely a) high-density cage (with the propellers closer to the center of mass, and b) expanded morphing arms when the system is carrying cargo and wants to fly more efficiently away from possible obstacles. 

\subsection{Morphing and Foldable CT-MAVs}

Morphing and foldable MAVs are a significant research direction with exciting results~\cite{floreano2017foldable,falanga2018foldable,bucki2019design}. In this manuscript we specifically focus on morphing and foldable designs that present and focus on collision-tolerance. The work in~\cite{patnaik2021collision} presents a quadrotor CT-MAV that integrates mechanical compliance based on a torsional spring such that when the outer arms collide, the impact time is increased and thus the effect on the main body is decreased. The authors analyze the post-collision dynamics and propose a recovery controller in order to stabilize the system to a hovering location. Safe collisions with speeds up to $2.5\textrm{m/s}$ are reported. Partially similar to this approach but much more tailored to navigation through narrow passages is SQUEEZE~\cite{patnaik2020design} which was previously presented in this manuscript and allows for shape reconfiguration during a collision. 

More miniaturized, the work in~\cite{dilaverouglu2020minicore} utilizes origami-like structures to make a foldable quadrotor with protective shrouds. In low-speed collisions (e.g., $1\textrm{m/s}$), the system enables navigation safety. Tailored to much larger systems with the ability to ferry significant payloads for cargo operations, the previously mentioned work in~\cite{kornatowski2020morphing} represents a foldable quadrotor design. When fully folded, the design offers a full collision-tolerant cage around its propellers and thus allows it to navigate cluttered environments (e.g., cities). When away from cluttered settings, it can unfold thus offering higher flight maneuverability but reduced collision protection. 

Not tailored to sustaining a collision and continuing its flight but primarily to morphing its shape and if needed sustaining a crash on the ground, the work in~\cite{desbiez2017x} utilizes two bicopter modules to form a reconfigurable X-configuration quadrotor. The system uses a magnetic and electrical joint between the quadrotor arms thus allowing for its easy removal and survivability against crashes. Results demonstrated that the design can decrease and increase its span dynamically mid-flight by up to $28\%$ within as little as $0.5\textrm{s}$. The prototype had a weight of $380\textrm{g}$. We conclude with pointing out the fact that several designs of foldable drones have shown exciting capabilities to access confined environments even though they have not explicitly focused on collision-tolerance~\cite{falanga2018foldable}. 

\subsection{Bimodal Aerial-Ground Designs}

A domain of research with high relevance to collision-tolerant MAVs is that of bimodal designs capable of both flight and ground locomotion. Such systems are tailored to be capable of traversing the ground and possibly present higher energetic efficiency as compared to when flying. Accordingly, they do present designs that allow for collision tolerance. 

One such design is the ``Drivecopter'' presented in~\cite{kalantari2020drivocopter}. The vehicle incorporates four independently actuated spherical wheels which beyond providing the means for ground mobility, they further protect the propellers and as opposed to passive wheels in other designs they allow to mitigate perceptual degradation in dusty environments that may be caused by propeller downwash. The system has a total weight of $5.1\textrm{kg}$ with the ground mobility elements contributing $0.9\textrm{kg}$, the aerial mobility elements $0.95\textrm{kg}$, the base chassis and power electronics $1.2\textrm{kg}$, the sensors $0.85\textrm{kg}$ and the battery $1.2\textrm{kg}$. The robot is designed having autonomy in mind and thus integrates a rich sensing payload. Similarly, the authors in~\cite{fan2019autonomous} also present a hybrid ground/aerial design albeit with simpler wheels and reduced protection around the propellers. The robot is also designed with autonomy in mind and integrates a VLP-16 3D LiDAR, a Realsense RGB-D camera, onboard light, a TeraRanger Evo 64px micro-camera, and an Intel NUC computing solution. Along the same research direction are the contributions in~\cite{tie2021,zhang2022autonomous}. In those, an autonomous aerial-terrestrial robot is presented and demonstrates $7$-fold increased energy efficiency when in terrestrial locomotion as compared to flying. With similar design, but at much smaller scale and without focus on sensor-based autonomous navigation, is the contribution in~\cite{pimentel2022bimodal}. 

Earlier work in the domain of bimodal aerial/ground designs, includes the ``HyTAQ'' system~\cite{kalantari2013design} which involves a quadrotor enclosed inside a rolling cage that is attached to the airframe by means of a shaft (fixed to the quadrotor) and a planar bearing (fixed to the rolling cage). Also presenting a rolling cage design but simultaneously offering the option to separate into two systems, the ``Shapeshifter''~\cite{agha2019shapeshifter} is a novel, multi-agent and bi-modal aerial/ground self-assembling robot for the exploration of Titan. The robot essentially allows separate aerial components to split and continue on their own. 

Focusing on enabling a flying robot to traverse very narrow passages, the authors in~\cite{fabris2021soft,fabris2022crash} presented a soft MAV that upon impact with a tight window/passage, compliantly squeezes its body and once it has entered the narrow passage proceeds by using tracks to crawl through. The design can traverse long passageways up to $34\%$ smaller than its nominal size. 

It is noted that a set of other contributions in the domain of bimodal aerial/ground systems have been proposed~\cite{jia2022quadrotor,lu2019design,atay2021spherical} reflecting the upcoming importance of this field. Among them we can also identify new trends with the contribution in~\cite{lu2019design} also containing limbs that can stretch out so that the robot may crawl on the supportive surface or facilitate carrying objects, landing on uneven ground and even walking. Finally, robots that perform wall-climbing by means of using propellers, such as the works in~\cite{beardsley2015vertigo,mahmood2021propeller}, are not discussed as flying is not considered to be one of their prime functionalities.

\subsection{Multi-linked reconfigurable CT-MAVs}

Multi-linked aerial robots represent a particular sub-category of flying machines with the core differentiating factor being that they are formed through the combination of a set of individual ``units''. Defining which systems exactly belong to this category is not necessarily clear but here we will focus on some notable examples that simultaneously present collision tolerance in their designs. 

The contribution in~\cite{oung2014distributed} presents the ``Distributed Flight Array (DFA)'' which is a flying platform that consists of multiple autonomous single-propeller vehicles that individually can merely drive, but when they dock with their peers they can also fly in a coordinated fashion. Each unit is in caged frame which beyond providing the surfaces for docking it also facilitates a means for collision-tolerance. 

The DFA is a distributed system but other solutions are centralized and involve continuously connected components. The authors in~\cite{zhao2018transformable} present a multirotor system with two-dimensional multi-links and is tailored to aerial transformation and manipulation, while each of its propellers comes with a protective shroud. The work in~\cite{zhao2018design} presented ``DRAGON'' which is a novel multi-linked system with each of its units integrating ducted fan propellers. The DRAGON prototype involved four links and demonstrated shape reconfiguration during flight. Relevant with these works is the ``Aerial Robotic Chain (ARC)''~\cite{nguyen2020reconfigurable,nguyen2021forceful,kulkarni2020reconfigurable} which involves multiple rigidly-caged quadrotors and has among others been used to fly through narrow passages and perform aerial manipulation tasks.

\subsection{A note on blimp-based designs}

This study has a limited scope in that it focuses on robotic systems that are tailored to miniaturization and have the ability to carry relatively significant payloads. Accordingly, we explicitly do not discuss designs that rely on lighter-than-air (e.g., blimp) structures to provide lift. However, it is noted that such systems indeed represent a class of collision-tolerant designs. Examples include the works in~\cite{song2020design,burri2013design,huang2019duckiefloat,palossi2019extending,gorjup2020low,troub2017simulation}.

\section{AUTONOMY FOR CT-MAVs}\label{sec:autonomy}

The majority of the research on CT-MAVs has focused on the mechatronic design aspects. There is in turn a limited set of contributions that either utilizes collision-tolerant designs that are equipped with advanced, albeit traditional navigation stacks, and a niche set of works that explicitly addresses how collision-tolerance can be exploited by and for autonomy. Below, we discuss a subset of such contributions with some of them also presented earlier in our design-based review.  

\subsection{Traditional autonomy-oriented CT-MAVs}

A set of works has considered the utilization of collision-tolerant designs combined with comprehensive autonomy stacks, analogous to those on systems without collision resilience, and aligned with the state of the art in the domains of Simultaneous Localization And Mapping (SLAM), as well as path and motion planning or learning-based navigation policies. 

The work in~\cite{rmfowl_icuas} presented RMF-Owl, one of the robots of Team CERBERUS made to explore the diverse underground environments of the DARPA Subterranean Challenge. The robot integrates a 3D LiDAR, vision, inertial sensors and an ARM-based computing solution which is capable of running simultaneously a) a multi-modal SLAM framework~\cite{khattak2020complementary}, b) a graph-based exploration path planning solution~\cite{kulkarni2022autonomous}, c) object-detection and localization based on YOLO v3~\cite{yolov3} and exploiting its co-integrated Neural Processing Unit, alongside d) full position control and other assisting functionalities. Similar sensing and processing payloads are found in a set of aerial/terrestrial bimodal designs~\cite{zhang2022autonomous,fan2019autonomous,kalantari2020drivocopter} with a subset of this research~\cite{fan2019autonomous,kalantari2020drivocopter} also relating to the efforts of Team CoSTAR~\cite{agha2021nebula}, one of other competing teams in the DARPA Subterranean Challenge. On a more miniaturized scale, TIERCEL~\cite{mulgaonkar2020tiercel} also aims for perception-driven autonomy albeit purely with vision sensors given its much reduced weight as compared to RMF-Owl. In between the size scale of RMF-Owl and TIERCEL we find the original RMF design~\cite{de2021resilient} which navigates purely based on visual-inertial methods and also integrates compliant flaps. Simultaneously, commercial systems with collision-tolerant designs are now making steps forward in integrating autonomy functionality such as localization and mapping or even path planning. 

Turning collision tolerance as an avenue to collect flight training data with relative safety, the work in~\cite{gandhi2017learning} used a Parrot Ar-Drone 2.0, which includes shrouds around its propellers, in order to collect data for learning a collision avoidance navigation policy without relying on few supervised flights or simulation. The authors crashed the robot $11,500$ times to create a rich dataset that --together with paths avoiding collisions-- enabled an efficient self-supervised policy to avoid collisions. 

The pertinent observation across these works is that the most advanced levels of perception-driven autonomy are observed on systems that employ relatively simple, mostly rigid, collision-tolerant designs with the focus being on integrating advanced but traditional sensing and processing solutions combined with sophisticated autonomy stacks. This is especially the case for purely flying systems (i.e., not bimodal systems which often come at much larger size and weight) such as RMF-Owl where every additional component of a more complex collision-tolerant design may further reduce the flight time of a system that already carries relatively heavy electronics. However, such an approach may limit collision tolerance capabilities both in terms of structural integrity but also in terms of the effects of impact to the core navigation functionalities (e.g., localization). 

\subsection{Contact-assisted Autonomy on CT-MAVs} 

A particularly interesting approach is taken by a set of contributions within which contacts are explicitly considered as part of the perception abilities of CT-MAVs. The authors in~\cite{lew2019contact} presents a method for Contact-Inertial Odometry (CIO). The work presents collision detection and estimation of odometry through collision, alongside reactive control and planning coupled with CIO. It uses estimated external forces to not only detect collisions but generate pseudo-measurements of the robot's velocity. It effectively allows a bimodal system to recover from possible failures of onboard localization and mapping that relies on exteroceptive sensors. 

Other works have  deeper focus on how collision-tolerance may be an avenue for autonomy with reduced perception capabilities. Examples include the works in~\cite{briod2013contact} where the previously discussed AirBurr system is used and a navigation framework that exploits sense of touch is developed. The robot makes use of eight miniature force sensors with each weighing as little as $0.9\textrm{g}$. More recently, the authors in~\cite{liu2021sensing} have also presented a design that utilizes the sense of touch. In particular, a quadrotor is combined with a suspended rim and at the central base the relative displacement of the rim - during a collision - is measured. As a more extreme concept, the work in~\cite{mulgaonkar2017robust} delivers robust swarmed flight with multiple micro CT-MAVs without collision avoidance. 

Further contributions that also exploit contact as part of their autonomy solutions include the previously discussed works on both rigid systems~\cite{mulgaonkar2020tiercel,mulgaonkar2017robust}, and elastic designs~\cite{briod2013contact,liu2021sensing}. Two relatively different ideas are considered by the authors in~\cite{khedekar2019contact} where a traditional quadrotor with propeller shrouds is enabled to navigate anomalous surfaces in contact, and the work in~\cite{wang2020fly} where a rigid caged robot can map a collision to a correct reactive maneuver to then avoid colliding again and coming back to a stable configuration. Finally, a much more detailed study on force estimation during contact is presented in~\cite{tomic2020simultaneous}, while more such contributions can be found in the community of aerial physical interaction and manipulation~\cite{ruggiero2018aerial}. The latter field includes designs that combine end effectors and collision-tolerant airframes such as the work in~\cite{jiang2023bridge}.

\subsection{Motion Planing for CT-MAVs}

CT-MAVs may employ motion planning methods applicable to aerial robots and treat collision tolerance as means to mitigate risks passively. However, a niche research direction relates to planning by accounting and exploiting collision resilience. The authors in~\cite{zha2021exploiting} build upon the RRT$^\star$ algorithm~\cite{lavalle1998rapidly,karaman2010optimal} and propose a method that as it detects collisions along a candidate motion primitive, it derives collision-inclusive trajectories for CT-MAVs to safely follow. Relevant to the above, the contribution in~\cite{RAMPLANNER_IROS_2022} considers how a motion planner can simultaneously account for localization uncertainty and the ability of a CT-MAV to afford collisions up to a certain impact kinetic energy in order to derive safe collision-inclusive paths that allow it to better navigate cluttered environments subject to uncertainty. Interestingly, relevant techniques are also applied for spacecraft systems~\cite{mote2020collision}. The work in~\cite{lu2021deformation} takes a different direction and presents trajectory replanning after a collision to estimate the post-impact state and derive a new waypoint to recover from the collision. The contribution in~\cite{mote2016framework,mote2017robotic} presents a Mixed Integer Programming (MIP) solution that identifies optimized trajectories which comprise certain planned collisions given a damage quantification model.

\section{DISCUSSION}\label{sec:discussion}

Research in collision-tolerant MAVs is a niche topic and accordingly a relatively small number of contributions have been presented. However, commercial success at least in the domain of confined indoor inspection alongside pioneering research contributions are clear indications of its importance. 

Perhaps a more significant indicator comes from a close look at the natural kingdom. The vast majority of flying species present the ability to tolerate collisions and this is an important property of their bodies. A look at honeybee wings reveals the important presence of resilin which is an elastomeric protein. The authors in~\cite{ma2015function} conduct a detailed study of the wings of honeybees and reveal that resilin stripes are found on both the dorsal and ventral side of the wings, while resilin patches were identified primarily on the ventral side. Its function is significant and among others is assumed to help in preventing wing damage during collisions. 

Supportive to the results of the above study is the earlier work in~\cite{mountcastle2014biomechanical}. Therein the authors focus specifically on the diverse biomechanical strategies for mitigating collision damage in insect wings. Studying yellowjacket wasps, they demonstrate the beneficial effect of a flexible resilin joint (the ``costal break'') positioned distally along the leading edge of their wing. This elastic joint allows the wing tip to crumple in a reversible manner when it collides with an obstacle. However, the authors also studied bumblebee wings where such a costal break was not found, yet significant wing damage was also not observed. This was explained by the different spatial arrangement of wing veins which support the wing. Relevant modeling studies were conducted and accordingly the different biomechanical strategies of embedding elastic materials or using different structural designs for the wings were explained. Both strategies may be particularly important for CT-MAVs and the different types of research, especially those of elastic, tensegrity-based, origami-based, and bioinspired designs. 

To understand the importance of safe collisions in flight we can also look at studies on birds. The work in~\cite{henningsson2021flying} studied birds flying through a wide range of gaps with widths ranging from larger than their wingspan down to just $1/4$ of their wingspan. It was shown that birds reduce their speed as they approach to fly through a tight gap, which in turn may reduce the likelihood of a collision or at least minimize its impact if it occurs. Relevant with this study is the contribution in~\cite{chin2022birds} where it is shown how birds can redirect forces by using their legs and wings not only to manoeuvre around obstacles but also make a controlled collision with a goal perch. Interestingly, studies indicate that birds repurpose the role of drag towards landing maneuvers~\cite{chin2019birds}. Simultaneously, the role of ankle elasticity and compliance during physical interaction is extensively studied for bird legs~\cite{hudson1936studies,roderick2021bird,rubenson2022running,zelik2014role,crandell2018coping}. The above observations, on how birds negotiate possible collisions and broadly physical interaction is particularly relevant to CT-MAV design and may provide cues towards new designs and algorithms. Among others, a key feature is the weight and inertia characteristics of natural flyers, insects or birds, with flying robots being mostly --by comparison-- heavier and presenting larger inertia. 

\section{DIRECTIONS FOR FUTURE RESEARCH}\label{sec:future}

The above studies and discussion allow us to identify directions of potentially fruitful and impactful research for collision-tolerant aerial robots. To define directions of future research, we first make three main observations:
\begin{itemize}
    \item Collision-tolerant designs are currently primarily studied with respect to the extent they can help in mitigating the risks of collisions in terms of ensuring the structural integrity of the vehicle. 
    \item Collision tolerance, especially by means of elastic and broadly energy-absorbing components, may not only help to ensure the structural integrity of a platform, but may also offer an avenue to reduce the degradation effects of collisions on the quality of navigation functionalities onboard a robot and especially those of localization/state estimation and control. 
    \item Collision tolerance is primarily thought as a last-resort mechanism to increase the survivability of a platform. Traditionally, such designs will be combined with means of collision avoidance and use their mechanical tolerance when a collision inadvertently happens. However, this separate treatment presents limitations as it does not explicitly account for a risk model of collisions and their autonomy-related effects. 
\end{itemize}
In view of the above, below we outline four selected directions of future research that may prove to be impactful. The discussed directions are not the only ones that one should consider but reflect our conclusions from the above study especially as they interplay with a focus on autonomous robotic systems.

\subsection{Ultra-lightweight Collision Tolerance}

Collision tolerance is an important property especially in cluttered environments. However, most of the designs indicate that it is associated with some non-negligible additional weight. This may not be of particular importance for robots that do not integrate significant sensing and processing payloads and thus maintain a high thrust(lift)-to-weight ratio and endurance. However, it can become significant for robots that are tailored to advanced operational autonomy and remote sensing tasks which are required to ferry significant payloads to execute their mission. Accordingly, it is important to work on the materials and design approaches that would ensure maximized collision-tolerance and impact absorption at a minimized weight. For example, where and how elasticity is introduced, where and when should origami-like structures be utilized, should we have overall protection with a cage or whether a more lightweight but almost-equally safe design can be achieved, are important questions and a design optimization process that also accounts for weight minimization is important.

\subsection{Bioinspired Flapping Wing CT-MAVs}

There is significant difference between the collision resilience in insects or birds and MAVs. Although it is not necessary that MAVs are always less collision-tolerant, most designs approach collision resilience in a manner different than flying species. One particular aspect of interest is the limited work on how flapping wing systems can retain flight control despite wing damage. It is possible that new wing designs, with elastic components or other means of energy absorption, and novel control policies, may offer enhanced collision resilience. Such a research direction may benefit from the progress in simulation tools, novel materials, and associated data-driven control policies for flapping wing MAVs~\cite{fei2019learning,fei2019flappy}.

\subsection{Proprioceptive Perception for Deformation Estimation} 

Morphing is an important avenue for achieving advanced collision tolerance as discussed. Particularly for the class of compliant CT-MAVs which are capable of exhibiting wider ranges of structural deformation, this may incur significant variability in their control allocation and perception (e.g., change in sensor-body transformation) policies, thus signifying the importance to account for it in the relevant loops. Explicit strain sensing at a number of sites within the robot structure can offer a first-degree solution to estimate the current deformation state, but not its dynamics --e.g., in impulse collision cases--. A compliance-aware structural model would allow such dynamic deformation estimation, but requires knowledge of collision contact points and magnitudes; self-evidently, a collision may occur at any site on an externally-facing component, thus making it inefficient to attempt to cover the CT-MAV with haptic devices. A novel alternative approach towards the goal of dynamic proprioceptive deformation estimation can be to leverage piezoacoustic sensing to determine the strike location and structural impact of a collision, using a small array of strategically placed acoustic transducers and the known acoustic propagation qualities of certain rigid parts composing the vehicle outer frame.


\subsection{Autonomy-oriented Collision Resilience}

Collision-tolerant designs should be studied and designed specifically with respect to their effect on essential autonomy functionalities. For example elastic collision-tolerant frames are likely to have a significant effect on both the proprioceptive --e.g., Inertial Measurement Unit (IMU)-- and exteroceptive --e.g., camera or LiDAR-- data captured by sensors utilized for onboard localization and state estimation (depending on their mounting, type, and configuration). The same holds for the interplay between collision-tolerant designs and their ability to retain control and come back to a stable configuration. To that end, rigid designs tend to maintain the same control allocation and thus their onboard control loops remain applicable post collision. Thus the ability to come back to a desired state simply depends on their particular design, stabilizability and tracking properties (e.g., on SO(3)/SE(3)). However, designs with elastic or other energy absorbing components may face a temporary change in their control allocation which should be explicitly considered, and despite this additional complexity they can be more resilient against collisions. If we model the interplay between collisions and such fundamental autonomy-enabling loops, it is possible that we will identify new designs that may offer enhanced properties with respect to the survivability of the robot both in terms of its structural integrity and its ability to maintain autonomous operation while relying purely on onboard sensing. 

\subsection{Holistic Treatment of Embodied Autonomy \& Resilience} 

If the above holds true, then a natural extension is to consider the advanced potential if embodied resilience by means of collision tolerance and autonomy resilience are co-designed. In the current view collision tolerance tends to be designed agnostically to autonomy loops, and autonomy functionalities do not account for how their data are affected during a collision. Limited works, such as those on risk-aware and collision-inclusive motion planning which were mentioned earlier attempt to account for this. However, advanced potential should arise if the problem is treated in a tight and holistic fashion. To that end we consider the possibility of using data-driven policy learning frameworks in which exteroceptive and proprioceptive sensor data, as well as estimation byproducts, before, during, and after a collision, are combined in order to derive an optimized autonomy pipeline that allows to exploit the robot's collision-tolerant design while navigating with safety and maximized efficiency with respect to its application domain. 

For the latter research direction, it should be noted that it can greatly benefit from high-fidelity simulation models for collision and physical interaction. This task can often be particularly complex. For example, compliant joints developed using materials such as silicone require more delicate Finite Element Method (FEM) analysis and cannot be accurately modeled with few discrete components. At the moment, design of such components and simulation of collision dynamics often takes place in high-fidelity multibody dynamics simulation environments allowing to simulate flexible components even subject to large motions and complex interactions with other structural components (e.g., using Adams~\cite{adamssimulator}). However, the research community has not developed, or at least not open-sourced, high-fidelity models for collision-tolerant aerial robots implemented in widely used simulators in robotics and especially those adopted for many software-in-the-loop and data-driven policy learning methods (e.g., Gazebo~\cite{gazebowebsite,koenig2004design} and the Open Dynamics Engine~\cite{smith2007open}, Isaac Sim~\cite{isaacsimwebsite,makoviychuk2021isaac} and PhysX~\cite{physxwebsite}, PyBullet~\cite{coumans2016pybullet}, OpenAI Gym~\cite{openaigym}, MuJoCo~\cite{todorov2012mujoco}, Webots~\cite{webotssim}, the Dynamic Animation and Robotics Toolkit (DART)~\cite{lee2018dart}, and others~\cite{korber2021comparing,liu2021role}). However, this imposes a limit in combined embodiment-and-autonomy research for collision-tolerant flying systems. It would be powerful to have the option to evaluate or train different navigation and overall autonomy solutions by taking into account accurate collision and interaction models. This in turn would allow to further exploit the interplay between collision-tolerance and autonomy, especially for systems involving energy-absorbing components, and identify new optima both in terms of collision-tolerant embodiments and collision resilience-aware autonomy for a given task. The evaluation of bioinspired versus more traditional rotorcraft designs should also be considered in that framework. 



\section{CONCLUSIONS}\label{sec:concl}

Collision-tolerant micro aerial vehicles represent a domain of research that can significantly contribute to the resilience of autonomous flying robots. Industrial adoption of collision-tolerant designs is already important, and this survey reveals that the domain represents a niche research field with rich results. Especially within cluttered environments or subject to degraded navigation conditions, collision tolerance may provide an avenue for enhanced resilience. In this work we reviewed key designs and outlined selected directions of future research especially as it relates to the interplay between collision tolerance and autonomy, of lightweight and bioinspired designs. Significant progress towards resilient flying robotic autonomy may be achieved if collision-tolerant embodiments and onboard navigation policies that exploit such airframe properties are co-designed. 




\section*{APPENDIX}
We have created, and plan to maintain, a website to list the literature in the domain. We have also included a form submission option in order to receive information for new relevant publications by authors or the community. It is available at~\url{https://s.ntnu.no/ct-mavs}.  






\bibliographystyle{IEEEtran}
\bibliography{SURVEY_CTMAV.bbl}

\end{document}